# Mixtures of Gaussians and Minimum Relative Entropy Techniques for Modeling Continuous Uncertainties

**William B. Poland**
poland@leland.stanford.edu

**Ross D. Shachter**
shachter@camis.stanford.edu

Department of Engineering-Economic Systems
Stanford University
Stanford, CA 94305-4025, USA

## Abstract

Problems of probabilistic inference and decision making under uncertainty commonly involve continuous random variables. Often these are discretized to a few points, to simplify assessments and computations. An alternative approximation is to fit analytically tractable continuous probability distributions. This approach has potential simplicity and accuracy advantages, especially if variables can be transformed first. This paper shows how a minimum relative entropy criterion can drive both transformation and fitting, illustrating with a power and logarithm family of transformations and mixtures of Gaussian (normal) distributions, which allow use of efficient influence diagram methods. The fitting procedure in this case is the well-known EM algorithm. The selection of the number of components in a fitted mixture distribution is automated with an objective that trades off accuracy and computational cost.

## 1 INTRODUCTION

Real probabilistic inference and decision analysis problems often involve continuous random variables (RVs) and decisions. Unfortunately, exact solution methods are not available for such problems in general, though several methods are available in the all-discrete case (Shachter 1986, Lauritzen and Spiegelhalter 1988, Pearl 1988, Jensen et al. 1990). Several approximate approaches are available, each with some disadvantages:

- The *discretization approach* quantizes the continuous distributions and uses a discrete solution method. Using a large number of points, as in numerical integration, is computationally burdensome, so practitioners often use only two or three points per variable. But representing continuous distributions accurately with a few points is tricky if the tails of the distributions are significant (Miller and Rice 1983, Keefer 1992). Also, discrete representations can hide simple continuous relationships, such as the linear conditional means in a multivariate Gaussian distribution.

- The *Monte Carlo approach* uses stochastic simulation of the RVs (*e.g.*, Shachter and Peot 1990). This sometimes requires a computationally burdensome number of simulations, even with techniques to speed convergence.

- The *moment approach* summarizes continuous distributions by their first few moments, either using these directly (*e.g.*, Howard 1971) or fitting discrete distributions fitted to the moments (Smith 1993). These methods can be inaccurate for irregular or multimodal distributions, and discretization has the disadvantages mentioned above.

- The *parametric approach* fits analytically tractable parametric distributions to the continuous RVs and an analytically tractable function to the utility function. An example is to fit a multivariate Gaussian (normal) distribution to the variables and a quadratic or concave exponential function of a quadratic to the utility function. The resulting model is easy to describe, assess, and solve in the notation of the Gaussian influence diagram, discussed below. However, the assumptions of the parametric approach often are overly restrictive.

This paper first outlines a parametric approach using mixtures of Gaussian distributions, which is much less restrictive than that of the Gaussian influence diagram. A reduction procedure generalizing Shachter and Kenley's Gaussian influence diagram procedure is proposed. For Bayesian updating, Lauritzen's message-passing procedure for conditional Gaussian models (Lauritzen 1992) should also be considered; it gives posterior probabilities for discrete variables and posterior means and variances for continuous variables. The main part of this paper gives procedures for transforming an arbitrary univariate distribution and fitting a mixture distribution to it, using the objective of minimizing relative entropy (Shore 1986). (Relative entropy is also known as Kullback-Leibler distance, directed divergence, cross-entropy, etc.) This objective results in the distribution of a desired form which maximizes the likelihood of a random draw from the input distribution. Multivariate generalizations of these procedures are mentioned.

The next section reviews influence diagrams and the Gaussian influence diagram case and generalizes this to a mixture-of-Gaussians influence diagram. Section 3 describes transformations to a univariate distribution to allow a more accurate Gaussian fit or a simpler mixture-of-Gaussians fit. Section 4 describes fitting a mixture to a (perhaps transformed) distribution, by the EM algorithm. Section 5 gives a procedure to automate selection of the number of components in a mixture, based on the EM algorithm. Finally, Section 6 describes some computational experience and gives directions for further research.

## 2 INFERENCE AND DECISION MAKING WITH MIXTURES OF GAUSSIANS: AN INFLUENCE DIAGRAM APPROACH

### 2.1 INFLUENCE DIAGRAMS

Influence diagrams (IDs) are useful for structuring and solving probabilistic inference problems and decision problems with a single decision maker (Howard and Matheson 1981). IDs are especially effective for representing finite mixture RVs, as will be seen in the next subsection. An ID is a directed acyclic graph in which the nodes represent the variables of the problem and the arcs represent probabilistic dependence or information. Oval nodes represent RVs, rectangular nodes represent decision variables, and a diamond (or rounded-rectangle or hexagonal) node represents the utility function expressing the decision maker's preferences. Sometimes the utility function can be expressed as a function of a single variable, called value, in which case often the utility node is omitted from the diagram and the value node is given the diamond shape.

To this standard notation, we add the placement of a number on the perimeter of any node to indicate a discrete variable with this number of values, or a "#" symbol on the perimeter to leave the number of values unspecified. The absence of a number or "#", as usual, does not restrict the variable to be continuous.

An arc from node X to node Y makes X a parent of Y and Y a child of X. There are two types of arcs. A conditioning arc, into a chance (RV) node, indicates that the child RV is conditioned on the parent variable. An informational arc, into a decision node, indicates that the value of the parent will be known at the time the decision is made. If a chance node has no parents, its distribution is marginal. A deterministic chance node, sometimes represented by a double oval, represents a variable for which the *conditional* distributions are all deterministic (though the marginal distribution might not be).

For a decision node, a decision policy specifies an alternative for each combination of outcomes of its parents (if any). An optimal decision policy maximizes expected utility, using the function in the utility node. A decision node may be thought of as a deterministic chance node for which the deterministic function, the decision policy, is initially unspecified.

The probabilistic inference problem—finding the probability distribution of some variables given others—can be solved by an ID procedure based on node removal and arc reversal operations that leave the joint distribution of the variables of interest unchanged (Shachter 1988, Shachter and Kenley 1989). Bayesian updating, or finding the posterior distribution of some variables when others have been observed, is then reduced to probabilistic inference followed by "instantiation" of the conditioning variables with the observations. After Bayesian updating, an ID with decisions can be solved for the optimal decision policies by the "reduction" procedure of Figure 1, based on similar operations and a decision optimization operation (Shachter 1986).

These inference and solution procedures are remarkably simple and efficient in the case of the Gaussian influence diagram (GID). A GID is an ID in which the joint probability distribution of all variables other than value or utility is multivariate Gaussian (so the decision domains are continuous and unbounded), and the utility function (if any) is a quadratic function of the other variables (the "value" function), or a concave exponential function of this (Shachter and Kenley 1989, Kenley 1986). The GID can be used to represent and solve well-known problems such as Kalman filters, linear-quadratic-Gaussian control problems, and linear regression problems. The simplicity of GID specification and operations follows largely from the following properties of the multivariate Gaussian distribution: (1) the conditional distribution of any variable given any others is Gaussian, (2) the conditional mean is a linear function of the conditioning variables, and (3) the conditional variance is constant (independent of the values of the conditioning variables).

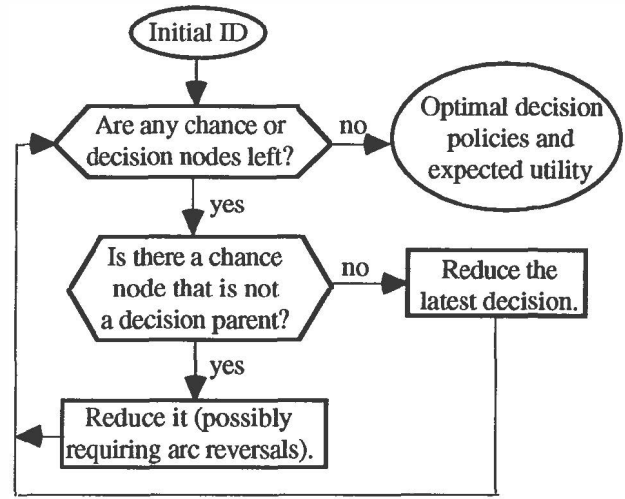

Figure 1: Reduction Procedure for Solving an ID

## 2.2 MIXTURE DISTRIBUTIONS AND THE MIXTURE-OF-GAUSSIANS INFLUENCE DIAGRAM

A finite mixture RV can be defined in terms of "component" and "selector" RVs as shown in Figure 2. "Finite" means the number of components is finite. A finite mixture RV is equal to one of its components; the selector variable selects which one but is unobserved. Figure 2b shows the selector, but not the components, separately from the mixture variable itself; an alternative partially expanded representation would show the components but not the selector separately. The probability distributions of the components are assumed to be of the same form, for example, Gaussian with mean and variance varying by component. Though the *distributions* of the selector and components are known, their *outcomes* are unknown, so the distribution of the mixture is an expectation, over the selector S, of the distributions of the components $X_1$ to $X_m$:

$$F_X(x) = \underset{S}{E} F_{X_S}(x) = p_1 F_{X_1}(x) + \ldots + p_m F_{X_m}(x),$$
$$f_X(x) = \underset{S}{E} f_{X_S}(x) = p_1 f_{X_1}(x) + \ldots + p_m f_{X_m}(x).$$

The component densities above are allowed to be Dirac delta functions, so that X can have probability mass at any point. Thus X can be mixed continuous and discrete, or as a special case, all-discrete.

A mixture RV should be distinguished from a weighted sum of the component RVs:

$$X \neq p_1 X_1 + \ldots + p_m X_m.$$

In the Gaussian-component case, for example, the latter is another Gaussian. One interpretation of a mixture variable is as the unknown, true model of a system or outcome of a process, where the components are the possible models/outcomes and the selector gives the probability of each being correct. Another interpretation is as a discrete variable with each outcome "blurred" by a random continuous perturbation. However, the component and selectors can be purely *artificial*, without any physical interpretation, and serve only to produce a desired distribution.

A mixture-of-Gaussians (MoG) RV or distribution is a mixture RV or distribution for which all components are Gaussian. A MoG distribution generalizes both a Gaussian distribution, which has only one component, and a discrete distribution, which has only zero-variance components. MoGs combine the simplicity of Gaussians and the flexibility of discrete distributions.

We define a MoGID as an ID such that after expanding any mixture-variable nodes as in Figure 2c:

1. There are no arcs from continuous to discrete nodes.
2. Dropping the discrete nodes (and any arcs from them) leaves a GID.

The MoGID could be represented as one parentless, combined discrete node with children in a GID.

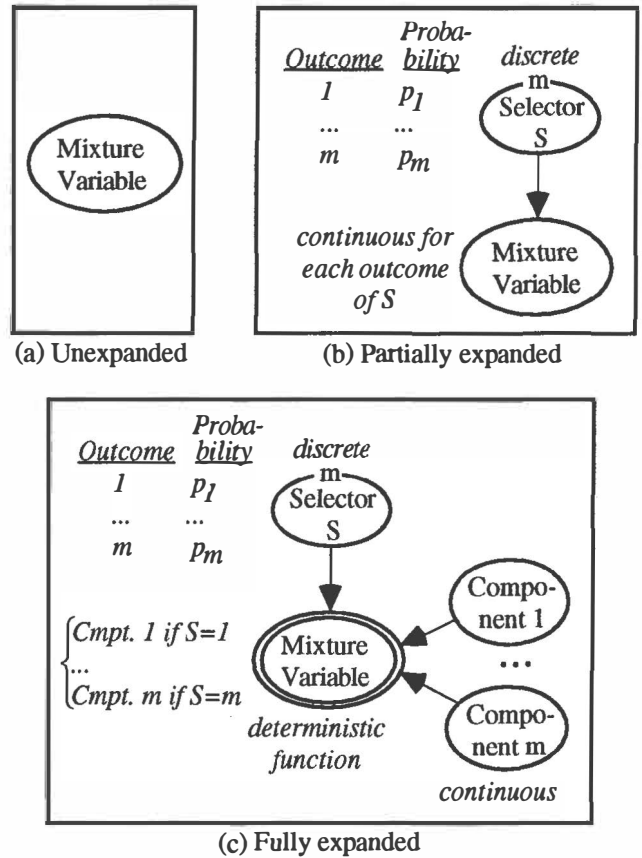

(a) Unexpanded   (b) Partially expanded

(c) Fully expanded

Figure 2: Influence Diagram Representations of a Mixture Variable. The selector "selects" a single (typically continuous) component for the mixture variable, but is unobserved, resulting in a convex combination of continuous distributions.

Alternatively, conditional dependence among MoGs in a MoGID could be represented by only arcs between selector nodes and other arcs between component nodes— though a simpler representation may be possible, as in the example below.

As an example, consider the oil wildcatter problem, a popular tutorial example in decision analysis posed by Raiffa (Raiffa 1968, p. xx):

> An oil wildcatter must decide whether or not to drill at a given site before his option expires. He is uncertain about many things: the cost of drilling, the extent of the oil or gas deposits at the site, the cost of raising the oil, and so forth. He has available the objective records of similar and not-quite-so-similar drillings in this same basin, and he has discussed the peculiar features of this particular deal with his geologist, his geophysicist, and his land agent. He can gain further relevant information (but still not perfect information) about the underlying geophysical structure at this site by conducting seismic soundings. This information, how-ever, is quite costly, and his problem is to decide whether or not to collect this information before he makes his final decision: to drill or not to drill.

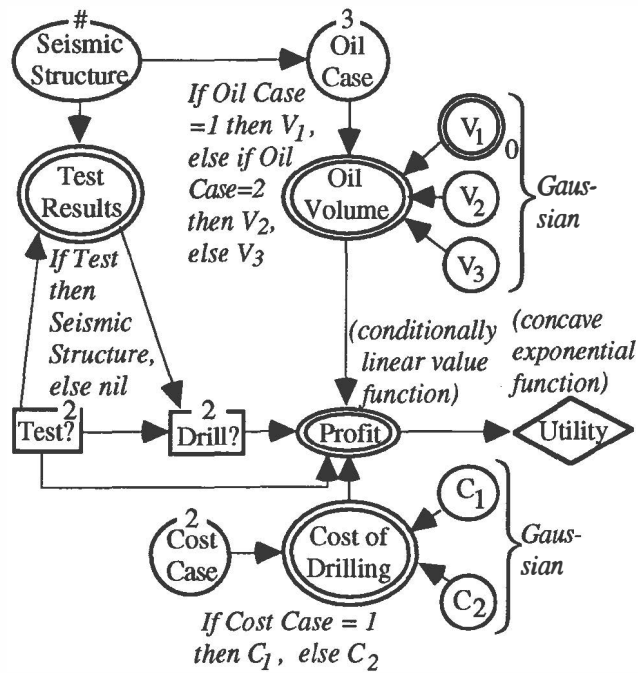

Figure 3: A MoGID Formulation of the Oil Wildcatter Problem. Simple artificial MoGs are fitted to assessed distributions for *Oil Volume* and *Cost of Drilling*.

Figure 3 shows a possible MoGID formulation of this problem, which recognizes that the *Oil Volume* RV is mixed continuous (for positive values) and discrete (at zero). The selector and components for both *Oil Volume* and *Cost of Drilling* are artificial, having been introduced only to allow tractable and reasonably faithful representations of the variables' distributions.

As long as arcs from continuous to discrete variables can be avoided, a MoGID can be solved by the reduction procedure of Figure 1, using discrete or Gaussian operations as appropriate and reducing continuous variables before discrete ones (Poland 1993). For the oil wildcatter problem, for example, the procedure removes nodes in a (non-unique) order such as: *Cost Of Drilling*, $C_1$, $C_2$, *Cost Case*, *Oil Volume*, $V_1$, $V_2$, $V_3$, *Oil Case*, *Seismic Structure* (first reversing the arc to *Test Results*), *Drill?*, *Test Results*, and *Test?*.

## 3 TRANSFORMATIONS TOWARD A DESIRED DISTRIBUTION

This section develops a procedure for transforming a univariate distribution to make it as close as possible, in a relative entropy sense, to a desired form, with emphasis on the Gaussian; the following two sections describe how to fit a MoG (or Gaussian) to the result of such a transformation. The entropy of a continuous RV X with density $f_X(\cdot)$ is

$$H(X) = -E[\ln f_X(X)] = -\int_{-\infty}^{\infty} f_X(x) \ln f_X(x)\, dx . \quad (1)$$

The relative entropy between two continuous RVs X and Y, with densities $f_X(\cdot)$ and $f_Y(\cdot)$ respectively, is

$$D(X,Y) = E\{ \ln[f_X(X)/f_Y(X)] \} . \quad (2)$$

Relative entropy is nonnegative, and zero only when the two densities are equal. Note also that

$$D(X,Y) \neq D(Y,X)$$

and that the support set of the second distribution must contain that of the first for the relative entropy to be finite. For convenience later, let

$$D_0(X,Y) = -E[\ln f_Y(X)] \quad (3)$$

so that relative entropy is a difference:

$$D(X,Y) = D_0(X,Y) - H(X) . \quad (4)$$

Consider a family of transformations $t_\lambda(X)$ parametrized by the vector $\lambda$, and consider an RV $Y(\theta)$ with a distribution of a desired form parametrized by the vector $\theta$. The transformation from the given family that minimizes relative entropy with a distribution of the desired form minimizes $D[t_\lambda(X), Y(\theta)]$ over both $\lambda$ and $\theta$.

It is convenient to minimize over $\theta$ first, holding $\lambda$ fixed. In particular, suppose the desired form is Gaussian. It can be shown that of all Gaussians, a continuous distribution has minimum relative entropy with the "moment-matching" Gaussian: the Gaussian with the same mean and variance. Therefore the transformation that minimizes the relative entropy with a Gaussian can be found by minimizing $D[t_\lambda(X), Y]$ over $\lambda$ only, where $Y \sim N(E[t_\lambda(X)], Var[t_\lambda(X)])$. This relative entropy can be expanded using expressions for the relative entropy of a variable with its moment-matching Gaussian and for the relative entropy of a differentiable monotonic transformation in terms of the untransformed variable. The result is

$$D[t_\lambda(X), Y] = 1/2 \{ 1 + \ln(2\pi) + \ln Var[t_\lambda(X)] \}$$
$$- H(X) - E[\ln |t_\lambda'(X)|]$$
where $Y \sim N(E[t_\lambda(X)], Var[t_\lambda(X)])$. \quad (5)

For example, consider the Box-Cox parametric family of "power/logarithm" transformations for positive variables (Box and Cox 1964):

$$t_p(x) = \begin{cases} \dfrac{x^p - 1}{p} & p \neq 0 \\ \ln x & p = 0 \end{cases} \quad \text{for } x > 0. \quad (6)$$

(For $p \neq 0$, the simpler transformation $x^p$ could have been used, but the scaled version above provides continuity at 0.) If X is a nonnegative RV, the distribution of $t_p(X)$ can be found by a change of variable:

$Y = t_p(X) \Rightarrow$
$f_Y(Y) =$
$\begin{cases} f_X[(py+1)^{1/p}] \, (py+1)^{1/p-1} & py+1 > 0, p \neq 0 \\ f_X(e^y) \, e^y & p = 0 \\ 0 & py+1 \leq 0 \end{cases}$ . (7)

Using $t_p(X)$ in (5) gives

$D[t_p(X), Y] = 1/2 \{ 1 + \ln(2\pi) + \ln \text{Var}[t_p(X)] \} - H(X) - (p-1) E[\ln X]$ where $Y \sim N(E[t_p(X)], \text{Var}[t_p(X)])$  (8)

so the power/logarithm transformation that has minimum relative entropy with a Gaussian can be found by a one-dimensional search (Hernandez and Johnson 1980):

$p^* = \arg\min_p D[t_p(X), Y]$
$= \arg\min_p \{ 1/2 \ln \text{Var}[t_p(X)] - (p-1) E[\ln X] \}$. (9)

If $p^*$ is close to a small integer or power that has a physical interpretation, it might be reasonable to round it to that number (and if $p \neq 0$, to substitute the simpler transformation $x^p$ at this time). Then the distribution of $t_{p^*}(x)$ can be found by (7).

The Box-Cox transformation family applies only to nonnegative RVs, and most naturally to ones with support on $[0,\infty)$. Others can be transformed in advance to make them nonnegative. A variable on $[a,b)$ can be transformed to one on $[0,\infty)$ by the "scaled odds" transformation $(x-a)/(b-x)$. For a variable on $(-\infty,\infty)$, it might be possible to find a practical upper and/or lower bound and use an affine or scaled odds transformation to make the support set $[0,\infty)$. This suggests the following transformation procedure for any RV: (i) transform it in advance to make it nonnegative, (ii) find $p^*$ from (9) and optionally round it, and (iii) find the transformed distribution with (7). Figure 4 shows the result of this procedure for exponential and uniform distributions. For the latter, first the support set is made $[0,\infty)$ by a scaled odds transformation; then the optimal power/logarithm transformation is found to be logarithmic. For further examples, see Hernandez and Johnson (1980).

Separate univariate transformations of dependent RVs may be undesirable if conditional dependence relationships among continuous variables have desired forms before transformation—for example, the linear conditional means and constant conditional variances required in the GID. One approach to this problem is to use the multivariate form of the power/logarithm transformation toward a Gaussian (Hernandez and Johnson 1980), which compromises between the objective of Gaussian marginal distributions and the objective of linear conditional means and constant conditional variances after transformation.

## 4  FITTING MIXTURE DISTRIBUTIONS WITH THE EM ALGORITHM

The EM (expectation, maximization) "algorithm" is an approach, rather than a single algorithm, for maximum-

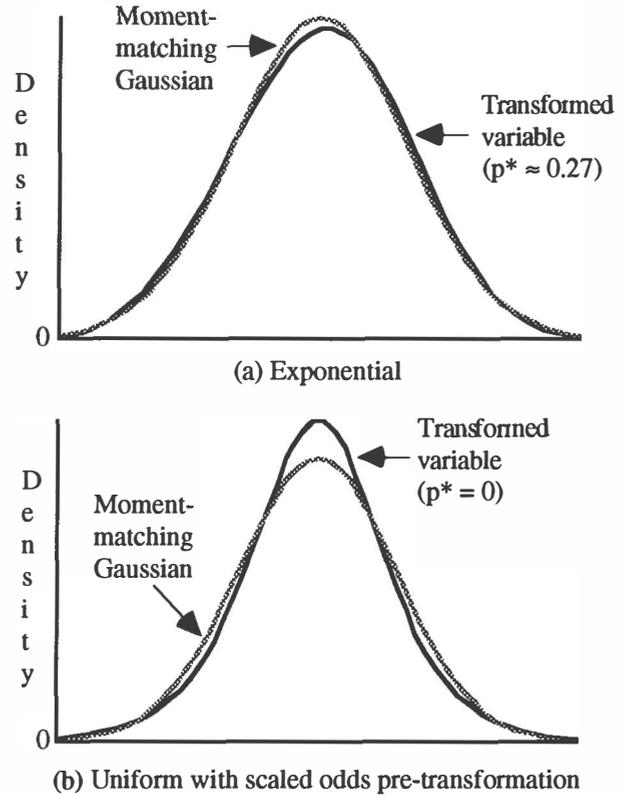

(a) Exponential

(b) Uniform with scaled odds pre-transformation

Figure 4: Some Densities after Power/Logarithm Transformations Toward a Gaussian. The results are close to the moment-matching (same mean and variance) Gaussians shown in gray.

likelihood estimation with incomplete data. Dempster *et al.* (1977) wrote the definitive paper on the approach, and Redner and Walker (1984) specialized it in detail to estimation of the parameters of mixture distributions, especially MoGs (with a given number of components). The usual input data for the EM algorithm is a sample of exchangeable observations. Since the purpose here is to fit a mixture distribution to a given *probability distribution*, rather than to given observations, a generalization of the EM algorithm is given which uses either kind of input data. Though only the univariate case is considered here, a multivariate generalization is straightforward and is interpreted in the final section.

Let X be an RV with a given distribution, and let $Y(\theta)$ be a mixture RV of a desired form (such as MoG), with m components and parameter vector $\theta = (p_1, ..., p_m, \theta_1, ..., \theta_m)$. For example, in a MoG, $\theta_i = (\mu_i, \sigma_i^2)$, the mean and variance of component i. Then the density of $Y(\theta)$ is

$f_Y(y \mid \theta) = p_1 \, f_1(y \mid \theta_1) + ... + p_m \, f_m(y \mid \theta_m)$

where $f_1(\cdot \mid \theta_1), ..., f_m(\cdot \mid \theta_m)$ are the component density functions. The objective considered here is to find the density of Y (not necessarily unique)—or equivalently, the value of $\theta$—that minimizes the relative entropy between X and Y over the feasible region $\Theta$ for $\theta$:

188 Poland and Shachter$$\begin{aligned}
\theta^* &= \arg\min_{\theta \in \Theta} D[X, Y(\theta)] \\
&= \arg\min_{\theta \in \Theta} D_0[X, Y(\theta)] \\
&= \arg\max_{\theta \in \Theta} E[\ln f_Y(X \mid \theta)] .
\end{aligned} \quad (10)$$

In the usual EM algorithm, the probability distribution of X is taken to be an *empirical* (frequency) distribution of a sample of exchangeable observations $\mathbf{x} = (x_1, ..., x_n)$ of draws $\mathbf{Y} = (Y_1, ..., Y_n)$ from Y. Thus $X = x_1$ or ... or $x_n$ and $D[X,Y(\theta)]$ is undefined, since X is discrete while $Y(\theta)$ is continuous. However, the definition of relative entropy for continuous variables, (2), can be generalized to apply to this discrete-with-continuous case by using (3) and (4) with X allowed to be discrete. With this generalized definition, since each of the n values of X has probability (frequency) 1/n,

$$\begin{aligned}
\theta^* &= \arg\max_{\theta \in \Theta} 1/n \sum_{j=1}^{n} \ln f_Y(x_j \mid \theta) \\
&= \arg\max_{\theta \in \Theta} \ln \prod_{j=1}^{n} f_Y(x_j \mid \theta) \\
&= \arg\max_{\theta \in \Theta} \ln f_Y(\mathbf{x} \mid \theta) \\
&= \arg\max_{\theta \in \Theta} f_Y(\mathbf{x} \mid \theta),
\end{aligned}$$

which is the maximum-likelihood estimate of $\theta$. Therefore the objective of minimizing the "relative entropy" between X and Y (extended to the discrete-with-continuous case) generalizes the objective of the EM algorithm: to maximize the likelihood of the observations $\mathbf{x}$ of draws from Y's distribution (Titterington *et al.* 1985, p. 115).

Since (10) cannot be solved in closed form, the EM algorithm iterates with a modification of it that replaces $\ln f_Y(X \mid \theta)$ by an expectation over the unknown selector associated with Y, S(Y), given the data X and the current parameter estimate $\hat{\theta}$:

$$\hat{\theta} \leftarrow \arg\max_{\theta \in \Theta} \underset{X}{E} \underset{S(Y)}{E} [\ln f_{Y,S(Y)}(Y,S(Y) \mid \theta) \mid Y = X, \hat{\theta}] . \quad (11)$$

When X is allowed to be an arbitrary RV (continuous or discrete), the EM algorithm has the same, attractive convergence properties as it does for the empirical distribution case, because the probability distribution of X can always be interpreted as a limiting case of a frequency distribution. (Convergence is discussed in Redner and Walker (1984). Techniques such as Aitken acceleration (Gerald and Wheatley 1984) are helpful to accelerate the typically slow convergence near the solution.)

The EM algorithm leads to the assignments

$$\hat{p}_i \leftarrow E\left[\frac{\hat{p}_i f_i(X \mid \hat{\theta}_i)}{f_Y(X \mid \hat{\theta})}\right], \quad i = 1, ..., m \quad (12)$$

and in the MoG case, with $\hat{\theta}_i = (\hat{\mu}_i, \hat{\sigma}_i^2)$,

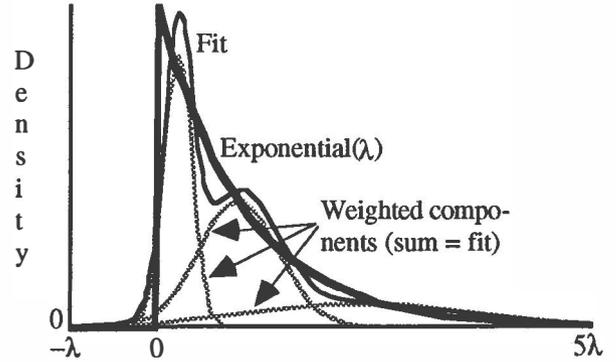

(a) Three-Component Fit to Exponential

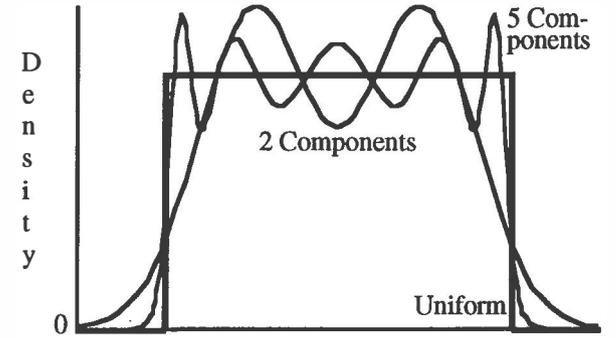

(b) Two and Five Component Fits to Uniform

Figure 5: Some MoG Fits by the EM Algorithm. MoGs with few components can approximate very non-Gaussian distributions.

$$\begin{aligned}
\hat{\mu}_i &\leftarrow E\left[\frac{X f_i(X \mid \hat{\theta}_i)}{f_Y(X \mid \hat{\theta})}\right] / E\left[\frac{f_i(X \mid \hat{\theta}_i)}{f_Y(X \mid \hat{\theta})}\right], \\
\hat{\sigma}_i^2 &\leftarrow E\left[\frac{(X - \hat{\mu}_i)^2 f_i(X \mid \hat{\theta}_i)}{f_Y(X \mid \hat{\theta})}\right] / E\left[\frac{f_i(X \mid \hat{\theta}_i)}{f_Y(X \mid \hat{\theta})}\right], \\
i &= 1, ..., m.
\end{aligned} \quad (13)$$

These assignments become equalities when the first-order necessary conditions for (10) are satisfied; in fact the EM algorithm for the MoG was developed without convergence theory from these conditions (Hasselblad 1966).

Figure 5 shows examples of MoGs fitted to exponential and uniform distributions (untransformed). Even without the help of transformations, densities far from Gaussian in shape can be fitted well with mixtures of a small number of components.

## 5 SELECTING THE MIXTURE SIZE

The EM algorithm provides the minimum-relative-entropy fit of a mixture distribution of a given "size," or number of components. How should this size be selected? If the distribution being fitted is known to be a finite mixture with components of the assumed form, the goal is to determine the true size. However, if the mixture serves as an *artificial* approximation of an arbitrary known distribution, a reasonable goal is to select the size that maximizes a utility function trading off preferences for accuracy and computational cost. An infinite mixture on $(-\infty, \infty)$ can replicate any distribution (for example, via an infinitude of zero-variance components along the desired distribution). On the other hand, the computational burden of calculations with mixtures rises with size at least as fast as for discrete distributions, which mixtures generalize.

Let $\theta_{(m)}$ be the parameter vector for a mixture of size m. If the probability distribution of X is the empirical distribution of exchangeable observations $x = (x_1, ..., x_n)$ of draws $Y = (Y_1, ..., Y_n)$ from a true mixture $Y = Y(\theta_{(m)})$, the likelihood for size m can be related to $D_0[X, Y(\theta_{(m)})]$:

$$\begin{aligned}
E[\ln f_Y(X \mid \theta_{(m)})] &= -D_0[X, Y(\theta_{(m)})] \\
1/n \sum_{j=1}^{n} \ln f_Y(x_j \mid \theta_{(m)}) &= -D_0[X, Y(\theta_{(m)})] \\
1/n \ln f_Y(x \mid \theta_{(m)}) &= -D_0[X, Y(\theta_{(m)})] \\
f_Y(x \mid \theta_{(m)}) &= \exp\{-n D_0[X, Y(\theta_{(m)})]\}.
\end{aligned} \quad (14)$$

If instead, an artificial mixture is to be fitted to a given distribution for X, the right-hand side of (14) is no longer a true likelihood, but it can be interpreted as an *accuracy measure* if n is interpreted as an "equivalent sample size," the size of an unspecified, hypothetical exchangeable sample that underlies the distribution of X. Dissociating n from the distribution of X this way is useful even in the empirical distribution case: as n increases, the EM algorithm need not be burdened by an increasingly detailed representation of the distribution of X. In practice, it may be possible to base n on an actual number of points, as when the distribution of X is found as a smooth curve drawn through n assessed points on a subjective cumulative distribution. For example, the software behind the interface in Figure 6, described below, uses a cubic spline through assessed cumulative points entered at the top of the window.

A convenient utility function for the artificial-mixture case is proportional to this accuracy measure and inversely proportional to the mixture size raised to a power k, as a computational cost measure. (Some statistical criteria for selecting model dimensionality, such as Akaike's information criterion and cross-validation, are similar in effect and could be substituted easily.) A possible setting for the parameter k is the total number of selectors in the probabilistic inference or decision problem; the cost measure is then a worst-case number of combinations of selector outcomes used in probability calculations if all mixtures have the same size. The maximum-utility size for this utility function can be estimated with the following heuristic: find the maximum-likelihood parameters for size m, $\theta^*_{(m)}$, with the EM algorithm, for m = 1, 2, ...; stop with size m when

$$\frac{f_Y(x \mid \theta^*_{(m+1)})}{(m+1)^k} < \frac{f_Y(x \mid \theta^*_{(m)})}{m^k},$$

or by (14) and (4), in terms of a relative entropy decrease and the combined parameter k/n, when

$$D[X, Y(\theta^*_{(m)})] - D[X, Y(\theta^*_{(m+1)})] < \frac{k}{n} \ln \frac{m+1}{m}. \quad (15)$$

(To reduce the risk of a local but not global maximum, one or more subsequent sizes could be checked too.) If k = 0 the search would not terminate, unless modified to incorporate prior probabilities for each size, resulting in a maximum *a posteriori* estimate (a geometric prior distribution on size is convenient; see Poland 1993). A similar search heuristic that decrements size is given in Cheeseman *et al.* (1988).

Figure 6 shows a user interface for fitting a MoG to an assessed cumulative distribution, with or without an automatic size search. It includes a "fast fit" option for a two-component mixture. For speed, this replaces the EM algorithm with the method of moments, which matches the first moments of the mixture and the input distribution (Cohen 1967). The results tend to diverge from those of EM as the input distribution diverges from a true mixture of two Gaussians. Unfortunately, the method of moments becomes intractable for larger mixture sizes.

## 6 CONCLUDING REMARKS

In experience with some theoretical distributions and empirical distributions from a semiconductor manufacturing process, power/logarithm transformations toward a Gaussian followed by EM fitting of a MoG with the size heuristic described above resulted in accurate fits requiring very few components. For a wide range of values of k/n in (15), the mixture size was typically greater than one but smaller than if the variable had not been transformed. The mixture size was one for the distributions of Figure 4.

It would be possible to nest inside the search over sizes a minimum-relative-entropy transformation toward a *mixture* of the current size. Whether this could be made computationally attractive is a research question. Another research direction is development of practical multivariate versions of the transformation and EM procedures, or less restrictive methods, to fit dependence relationships in an analytically convenient way. The multivariate EM algorithm can be thought of as fitting multiple univariate mixtures with a common selector; multiple dependent selectors would provide more generality while allowing efficient calculations in a MoGID. Also, research is needed to generalize existing models that exploit the tractability of Gaussians but suffer from their restrictions. Simple MoGs might provide valuable flexibility at an acceptable computational price.

### Acknowledgments

This research was supported in part by a gift from Intel Corporation and benefited from discussions with our colleagues in the Engineering-Economic Systems Department.

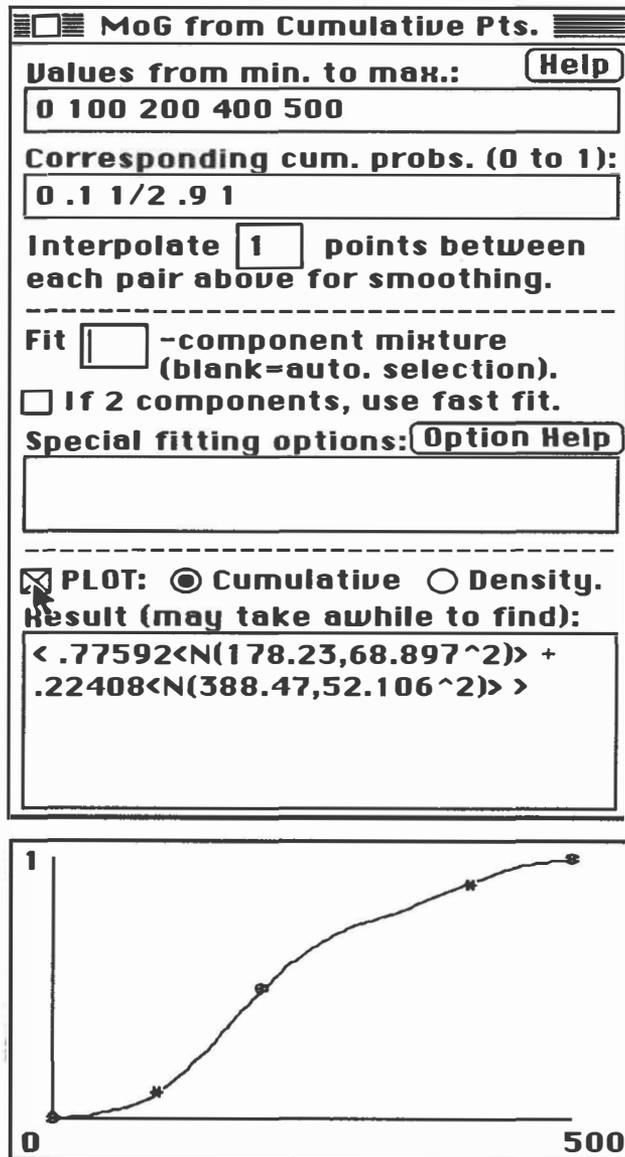

Figure 6: A User Interface for Fitting a MoG to an Assessed Distribution. The graph shows a 2-component MoG fitted to a distribution passing through 5 input cumulative points. The fit is slightly low at the middle point.

## References

Box, G. E. P. and D. R. Cox (1964). "An Analysis of Transformations." *J. Royal Statist. Soc.*, B, 26, 211-243; discussion: 244-252.

Cheeseman, P., J. Kelly, M. Self, J. Stutz, W. Taylor, and D. Freeman (1988). "AutoClass: A Bayesian Classification System." *Proc. Fifth Intl. Conf. on Machine Learning*, University of Michigan, Ann Arbor.

Cohen, A. C. (1967). "Estimation in Mixtures of Two Normal Distributions." *Technometrics*, 9, 15-28.

Dempster, A. P., N. M. Laird, and D. B. Rubin (1977). "Maximum Likelihood Estimation from Incomplete Data via the EM Algorithm." *J. Royal Statist. Soc.*, B, 39, 1-38.

Gerald, C. E. and P. O. Wheatley (1984). *Applied Numerical Analysis*. Reading, MA: Addison-Wesley.

Hasselblad, V. (1966). "Estimation of Parameters for a Mixture of Normal Distributions." *Technometrics*, 8, 3, 431-444.

Hernandez, F. and R. A. Johnson (1980). "The Large-Sample Behavior of Transformations to Normality." *J. Am. Statist. Ass.*, 75, 372, 855-861.

Howard, R. A. (1971). "Proximal Decision Analysis." In Howard, R. A. and J. E. Matheson (Eds.), *Readings on the Principles and Applications of Decision Analysis*. Vol. II. Menlo Park, CA: Strategic Decisions Group, 1984.

Howard, R. A. and J. E. Matheson (1981). "Influence Diagrams." In Howard, R. A. and J. E. Matheson (Eds.), *The Principles and Applications of Decision Analysis*. Vol. II. Menlo Park, CA: Strategic Decisions Group, 1984.

Keefer, D. L. (1992). "Certainty Equivalents for Three-Point Discrete-Distribution Approximations." Working paper, Department of Decision and Information Systems, Arizona State University, Tempe, AZ.

Kenley, C. R. (1986). "Influence Diagram Models with Continuous Variables." Ph.D. Dissertation, Department of Engineering-Economic Systems, Stanford University, Stanford, CA.

Jensen, F. V., K. G. Olesen, and S. K. Andersen (1990). "An Algebra of Bayesian Belief Universes for Knowledge-Based Systems." *Networks*, 20, 5, 637-659.

Lauritzen, S. L. (1992). "Propagation of Probabilities, Means, and Variances in Mixed Graphical Association Models." *J. Am. Statist. Ass.*, 87, 420, 1098-1108.

Lauritzen, S. L. and D. J. Spiegelhalter (1988). "Local Computations with Probabilities on Graphical Structures and their Application to Expert Systems." *J. Royal Statist. Soc.*, B, 50, 2, 157-224.

Miller, A. C. and T. R. Rice (1983). "Discrete Approximations of Probability Distributions." *Mgmt. Sci.*, 29, 352-362.

Pearl, J. (1988). *Probabilistic Reasoning in Intelligent Systems*. San Mateo, CA: Morgan Kaufmann.

Poland, W. B. (1993). "Decision Analysis with Continuous and Discrete Variables: a Mixture Distribution Approach." Ph.D. Dissertation, Department of Engineering-Economic Systems, Stanford University, Stanford, CA, forthcoming.

Raiffa, H. (1968). *Decision Analysis*. Menlo Park, CA: Addison-Wesley.

Redner, R. A. and H. F. Walker (1984). "Mixture Densities, Maximum Likelihood, and the EM Algorithm." *SIAM Rev.*, 26, 2, 195-239.

Shachter, R. D. (1986). "Evaluating Influence Diagrams." *Opns. Res.*, 34, 871-882.

Shachter, R. D. (1988). "Probabilistic Inference and Influence Diagrams." *Opns. Res.*, 36, 589-604.

Shachter, R. D. and C. R. Kenley (1989). "Gaussian Influence Diagrams." *Mgmt. Sci.*, 35, 5, 527-550.

Shachter, R. D. and M. A. Peot (1990). "Simulation Approaches to General Probabilistic Inference on Belief Networks." In Henrion, M. *et al.* (Eds.), *Uncertainty in Artificial Intelligence 5*. New York: North-Holland,.

Shore, J. E. (1986). "Relative Entropy, Probabilistic Inference, and AI." In Kanal, L. N. and J. F. Lemmer (Eds.), *Uncertainty in Artificial Intelligence*. New York: North-Holland.

Smith, J. E. (1993). "Moment Methods for Decision Analysis." *Mgmt. Sci.*, 39, 3, 340-358.

Titterington, D. M., A. F. M. Smith, and U. E. Makov (1985). *Statistical Analysis of Finite Mixture Distributions*. New York: Wiley.